%% file: root.tex
\documentclass[letterpaper, 10 pt, conference]{ieeeconf}  

\IEEEoverridecommandlockouts                              

\usepackage[OT1]{fontenc}

\usepackage{amsmath} 

\usepackage{graphicx}
\usepackage{sigmund_macros}
\usepackage{multirow}
\usepackage{soul}
\usepackage[frozencache,cachedir=.]{minted}
\usepackage{listings}
\usepackage{todonotes}
\usepackage{svg}
\usepackage{amsmath}
\usepackage{amssymb}
\usepackage{booktabs}

\usepackage{algorithm}
\usepackage{algorithmic}

\usepackage{subcaption}
\usepackage{adjustbox}

\usepackage{pgfplots}
\pgfplotsset{compat=1.17}
\usepackage{xspace}
\usepackage{url}

\usepackage{bibentry}
\definecolor{deeppink}{rgb}{1.0, 0.08, 0.58}

\usepackage{xcolor}
\usepackage{url}
\usepackage{hyperref}
\hypersetup{
  colorlinks = true,  
  urlcolor   = deeppink,  
  linkcolor  = blue,  
  citecolor  = red,  
  pdfview    = XYZ,  
}

\allowdisplaybreaks

\usepackage{hyperxmp}

\title{\LARGE \bf
Fast Policy Synthesis with Variable Noise Diffusion Models 
}

\author{Sigmund H. Høeg$^{1}$, Yilun Du$^{2}$ and Olav Egeland$^{1}$
\thanks{$^{1}$Department of Mechanical and Industrial Engineering,
        Norwegian University of Science and Technology (NTNU)
        {\tt\small \{sigmund.hoeg, olav.egeland\}@ntnu.no}}%
\thanks{$^{2}$ Harvard University,
        {\tt\small ydu@seas.harvard.edu}}%
}
\begin{document}

\maketitle
\thispagestyle{empty}
\pagestyle{empty}

\begin{abstract}

Diffusion models have seen rapid adoption in robotic imitation learning, enabling autonomous execution of complex dexterous tasks. However, action synthesis is often slow, requiring many steps of iterative denoising, limiting the extent to which models can be used in tasks that require fast reactive policies. To sidestep this, recent works have explored how the distillation of the diffusion process can be used to accelerate policy synthesis. However, distillation is computationally expensive and can hurt both the accuracy and diversity of synthesized actions. We propose \model (\modellong), an alternative method to accelerate policy synthesis, leveraging the insight that generating a partially denoised action trajectory is substantially faster than a full output action trajectory. At each observation, our approach outputs a partially denoised action trajectory {\it with variable levels of noise corruption}, where the immediate action to execute is noise-free, with subsequent actions having increasing levels of noise and uncertainty. The partially denoised action trajectory for a new observation can then be quickly generated by applying a few steps of denoising to the previously predicted noisy action trajectory (rolled over by one timestep).  We illustrate the efficacy of this approach, dramatically speeding up policy synthesis while preserving performance across both simulated and real-world settings. Project website: \href{https://streaming-diffusion-policy.github.io}{https://streaming-diffusion-policy.github.io}.

\end{abstract}

\input{Sections/intro}

\input{Sections/related_work}

\input{Sections/Method/method}

\input{Sections/Experiments/experiments}

\input{Sections/conclusion}

\IEEEtriggeratref{39}





\clearpage

\bibliographystyle{IEEEtran}
\bibliography{references}


\end{document}

%% file: Sections/intro.tex
\section{Introduction}
\label{sec:Introduction}

Diffusion models have proven to be powerful tools for robotic imitation learning, and are able to express complex and multimodal distributions over action trajectories \cite{chiDiffusionPolicyVisuomotor2023b,reussGoalConditionedImitationLearning2023a}. However, diffusion models often require multiple iterations to obtain a clean prediction, where each iteration requires evaluating a neural network, which is typically large in size. This limits the application of diffusion-based policies to environments with fast dynamics that require fast control frequencies, restricting them to more static tasks, such as pick-and-place operations. In addition, the scarcity of computational resources onboard mobile robots further motivates the need to minimize the computation required to predict actions using diffusion-based policies.

Recent techniques have focused on reducing the number of steps necessary to predict actions through distillation \cite{kim2024consistency, prasadConsistencyPolicyAccelerated2024}, but such a procedure can often be expensive and unstable and can further hurt both the quality and diversity of synthesized samples \cite{meng2023distillation}. In addition, distilled models lose useful properties of diffusion process such as composability \cite{du2023reduce, wang2024poco} and approximate likelihood computation \cite{zhou2024adaptive}.

\begin{figure}[t]
    \centering
    \includegraphics[width=\linewidth]{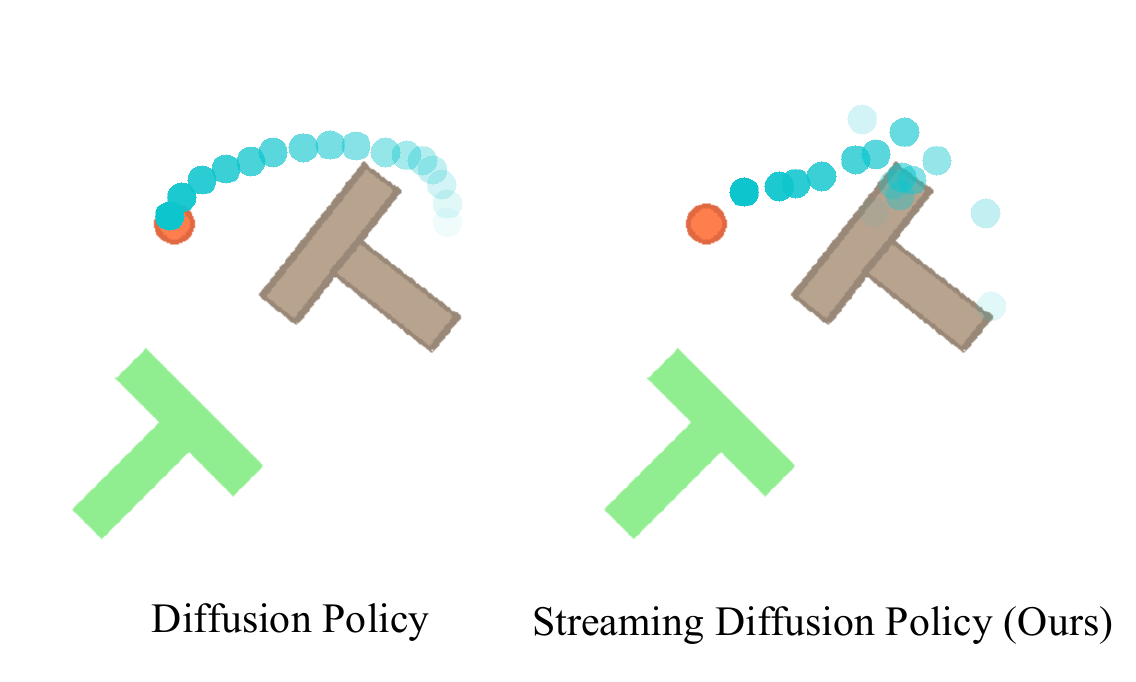}
    \caption{\textbf{Predicting Partially Denoised Trajectories} Previous applications of diffusion models for robotic control, like Diffusion Policy~\cite{chiDiffusionPolicyVisuomotor2023b} plans over a longer horizon, and executes the first few steps, discarding the rest after sampling. \modellong, however, denoise only the first actions, and keeps the future actions partly denoised for future predictions.}
    \vspace{-20pt}
    \label{fig:sampling_speed}
\end{figure}

In this paper, we propose an alternative approach to significantly speed up the synthesis of actions using diffusion models, while maintaining the simplicity and properties of a typical diffusion model. Our key insight is that robot execution only requires access to the immediate action we wish to execute. Therefore, instead of generating a full action trajectory to execute, our approach, \modellong (\model) generates a partially denoised action trajectory, where the immediate next action to execute is noise-free, with subsequent actions in the trajectory having increasing levels of noise and uncertainty (with the last action being close to pure noise). We illustrate this in Figure~\ref{fig:sampling_speed}.

This choice of generating partially denoised trajectories of increasing uncertainty enables us to substantially accelerate policy synthesis. Given a partially denoised action trajectory generated from the previous observation, we can execute the first action in this trajectory, remove this action from the trajectory, and then add a Gaussian noise action at the end of the trajectory. This forms a new action trajectory, where every action in the trajectory has an increased noise level. By running a few steps of denoising on this trajectory given the current observation, we can recover a new partial denoised action trajectory that we can execute at the observed timestep. By recursively applying this procedure given each observation, we can substantially reduce the number of denoising iterations needed at each observation to generate actions.

In practice, at the start of policy execution, we do not have a partially generated action trajectory to initialize policy synthesis. We explore a set of methods to initialize these action trajectories and find that a simple zero initialization of actions leads to effective performance. In addition, to adapt our approach to settings with very high control frequencies, where we do not have time to run network inferences at each step of action, we introduce a form of \model with {\it action chunking}. In this setting, a chunk of actions is generated at once, which can be executed using open-loop control while we use our recursive procedure to generate the next chunk of actions to execute.

Overall, our contributions in this paper are three-fold. First, we introduce a \model, an approach to accelerate visuomotor policy synthesis by generating a partially denoised action trajectory with variable levels of noise at each action. We then explore various noise corruption schemes to apply during training time and explore different sampling schemes for initializing and implementing recursive sampling of action sequences. Finally, we show the efficacy of this approach improving speed with little impact on performance across both simulated and real-world domains.

%% file: Sections/related_work.tex
\section{Related Work}
\label{sec:Related Work}

Diffusion models can express complex multimodal distributions and show stable training dynamics and hyperparameter robustness. This has resulted in a wide application in motion planning \cite{jannerPlanningDiffusionFlexible2022a, luo2024potential, carvalho2023mpd, 10610519, huang2023diffusion}, imitation learning \cite{pearceImitatingHumanBehaviour2022,chiDiffusionPolicyVisuomotor2023b, pmlr-v229-ha23a, pmlr-v229-xian23a, liCrosswayDiffusionImproving2024,Ze2024DP3,liCrosswayDiffusionImproving2024, Wang-RSS-24, Chen-RSS-24, sridhar2023memoryconsistent, zhao2024aloha, Chi-RSS-24}, goal-conditioned imitation learning \cite{reussGoalConditionedImitationLearning2023a, Reuss-RSS-24, chen2023playfusion, language-control-diffusion}, and grasp prediction \cite{urain2022se3dif}. Most of these works focus on sequential trajectory generation by denoising over the full horizon; for instance, Diffuser~\cite{jannerPlanningDiffusionFlexible2022a} produces a sequence spanning the entire episode while Diffusion Policy~\cite{chiDiffusionPolicyVisuomotor2023b} samples plans over a shorter action horizon. However, these methods can be computationally slow when applied sequentially in real-world environments. In contrast, our streaming approach can be integrated with these existing methods to accelerate the decision-making process significantly while maintaining the benefits of diffusion-based planning.


While there exist many examples of methods speeding up the prediction for image-based diffusion models \cite{karrasElucidatingDesignSpace2022a, songConsistencyModels2023a,songDenoisingDiffusionImplicit2020, kim2024consistency} there are also works that have looked at speeding up the prediction specifically for diffusion-based robotic policies. Reuss et al.~\cite{reussGoalConditionedImitationLearning2023a} optimize a diffusion model for goal-conditioned action generation with 3 denoising steps by choosing a suitable training and sampling algorithm. Their contribution are orthogonal to our method, and applying a sampling scheme akin to \model to their framework can decrease the sampling time further. Recently, Consistency Policy by \cite{prasadConsistencyPolicyAccelerated2024} shows that Consistency Trajectory Models \cite{kim2024consistency}, originally designed for speeding up image generation models, also perform comparably to Diffusion Policy while being a magnitude faster. They find that a 3-step method results in higher performance than the 1-step method, especially on tasks requiring long-horizon planning. This multi-step approach can also be combined with sampling framework of \model to increase the sampling speed. Another direction extends the work on Dynamical Motion Primitives (DMPs), as exemplified by \cite{Scheikl2024MPD}, to increase the sampling speed. However, introducing a formulation based on DMPs introduce some restrictions, such as producing smoother trajectories than the demonstrations, which might not be suitable for all tasks. 

Closely related to our approach are Rolling Diffusion~\cite{pmlr-v235-ruhe24a} and Temporally Entangled Diffusion~\cite{zhangTEDiTemporallyEntangledDiffusion2023}, both of which consider the rolling-window approach to producing sequences of data with diffusion models. The former applies the framework to video prediction and fluid dynamics forecasting, and the latter to character motion generation. Diffusion Forcing~\cite{chen2024diffusionforcingnexttokenprediction} also allows for an independent noise level per token in the sequence and applies it to sequence prediction such as motion planning, video prediction, and decision making. AR-Diffusion~\cite{10.5555/3666122.3667859} applies a linearly increasing noise level for language modeling. Unlike these works, we consider the setting of a closed-loop robotic policy.


%% file: Sections/Method/method.tex
\input{Sections/Method/Subsections/intro}
\input{Sections/Method/Subsections/buffer}
\input{Sections/Method/Subsections/sampling}
\input{Sections/Method/Subsections/training}

%% file: Sections/Method/Subsections/intro.tex
\section{\modellong}

A visuomotor policy solves the task of observing a sequence of observations $\mathbf{O}_t$ of length $T_o$ and predicts the next action $\mathbf{a}_t \in \mathbb{R}^{D_a}$ to apply in the environment. Diffusion Policy \cite{chiDiffusionPolicyVisuomotor2023b} solves this by modeling the conditional distribution $p(\mathbf{A}_t | \mathbf{O}_t)$ of an action sequence $\mathbf{A}_t = [\mathbf{a}_t,\ldots,\mathbf{a}_{t+T_a}]$ of length $T_a$ into the future. However, only the first action (or few actions in the case of open-loop control) in $\mathbf{A}_t$ are executed before re-planning, effectively discarding the excess steps.
In contrast, our \model approach synthesizes a persistent \textit{action buffer} consisting of future actions to execute. We iteratively update this buffer, which we detail in Subsection \ref{sec:Action Buffer}. The sampling process is described in Subsection \ref{sec:Sampling from TEDi Policy} and training details in Subsection \ref{sec:Training TEDi Policy}.

%% file: Sections/Method/Subsections/buffer.tex
\begin{figure}
    \centering
    \hfill
    \vspace{0.5em}
	
	\includegraphics[width=0.95\linewidth]{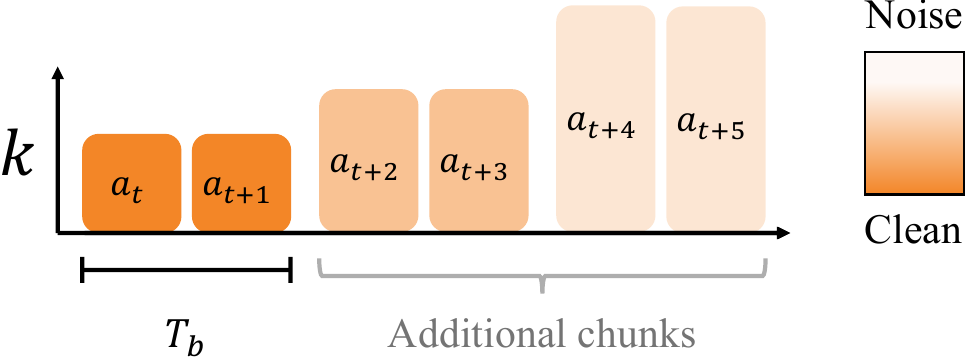}
	\caption{\textbf{Action Buffer Visualization.} \model keeps a persistent \textit{action buffer}, assigning actions at the beginning of the buffer to low noise levels and future actions to higher noise levels. This reduces denoising iterations for future action synthesis.}
	\label{fig:TEDi_Buffer}
 \vspace{-15pt}
\end{figure}
\subsection{Action Buffer}
\label{sec:Action Buffer}
Similar to Diffusion Policy \cite{chiDiffusionPolicyVisuomotor2023b} we denoise an clean action sequence $\mathbf{A}_t$ of $T_a$ future actions. However, in contrast to Diffusion Policy, our proposed method \model denoises each future action in $\mathbf{A}_t$ with a different level of increasing noise. Actions outside the immediate action that is executed are stored in {\it action buffer} that is reused to generate actions in future observations. 

In Diffusion Policy \cite{chiDiffusionPolicyVisuomotor2023b}, the whole action sequence is denoised in $N$ steps by decreasing the diffusion step $k$:
\begin{equation}
    \mathbf{A}_t^{k-1} \gets \alpha_k(\mathbf{A}_t^k - \gamma_k \epsilon_\theta(\mathbf{A}_t^k;\mathbf{O}_t, k) + \mathcal{N}(0, \sigma_k^2)).
\end{equation}
In \model, the action $\mathbf{A}_{t,i}$ at action timestep $i$ is denoised at a separate noise level $k_i$ from $\mathbf{k} = [k_0, k_1, \dots, k_{T_a-1}]$: 
\begin{equation}
    \mathbf{A}^{\mathbf{k}-1}_{t} \gets \alpha_{\mathbf{k}}(\mathbf{A}^{\mathbf{k}}_{t} - \gamma_{\mathbf{k}} \epsilon_\theta(\mathbf{A}_{t}^{\mathbf{k}};\mathbf{O}_t, \mathbf{k}) + \mathcal{N}(0, \sigma_{\mathbf{k}}^2)),
\end{equation}
with $\alpha_{\mathbf{k}}=[\alpha_{k_0}, \alpha_{k_1}, \dots, \alpha_{k_{T_a-1}}]$. The coefficients $\gamma_{\mathbf{k}}, \sigma_{\mathbf{k}}$ are constructed similarly. Each action starts at a diffusion level of $k_i = N$ corresponding to pure noise with $k_i = 0$ corresponding to a clean action. We store future actions $\mathbf{A}_{t,i}$ for $i > 1$ in an action buffer that is used to generate actions in future observations (illustrated in Figure~\ref{fig:TEDi_Buffer}).

To enable fast control rates, in \model, we divide noise levels across actions into chunks  \cite{zhaoLearningFineGrainedBimanual2023}, where we group the $T_a$ steps in the buffer into $h$ \emph{chunks} $\mathbf{A}_{t,i}$ of length $T_b$ where $i = 0,\ldots,h-1$. This enables $T_b$ actions to be generated at once, allowing these actions to be immediately executed while subsequent actions are generated. 


%% file: Sections/Method/Subsections/sampling.tex
\subsection{\model Policy Sampling}
\label{sec:Sampling from TEDi Policy}

\input{Figures/tedi_sampling_and_noise_injection_schemes}

The choice of storing an action buffer with increasing levels allows us to substantially increase sampling speed in \model. We illustrate how this enables fast sampling and how we can initialize the action buffer.

\subsubsection{Fast Sampling with Action Buffers} 
\label{sect:sampling_chunkwise}
The use of an action buffer described in the previous section allows us to implement fast recursive sampling. The recursive sampling process can be divided into two main steps: (a) lightly denoising the action buffer to get the next action to execute (b) applying the first chunk to the environment and appending noise to refresh the buffer. We provide pseudocode for the sampling procedure in Algorithm \ref{alg:tedi_sampling}. 

\textbf{Denoising.} The action buffer contains a trajectory of actions where each action chunk is associated with increasing noise levels 
\begin{equation}
    \left[ \frac{N}{h}, \frac{2N}{h}, \dots, N \right].
\end{equation}
By denoising the buffer for $\frac{N}{h}$ steps, the first chunk will be denoised and marked with the diffusion level $k_{i=0}=0$. The diffusion levels for each chunk are then
\begin{equation}\left[0,  \frac{N}{h}, \dots, \frac{N(1-h)}{h}\right].\end{equation}

\textbf{Removing first chunk and appending noise.} Actions in the first chunk are now noise-free and applied to the environment. Next, we add a new last chunk that is initialized by pure noise. This results in a buffer with noise levels 
\begin{equation}
    \left[ \frac{N}{h}, \frac{2N}{h}, \dots, N \right],
\end{equation}
which is equivalent to the noise levels at the beginning of sampling. This enables us to apply the same fast denoising procedure to obtain new actions to execute given new observations. Note that the degree of action generation is accelerated is inversely proportional to the number of chunks $h$, as we run a total of $\frac{N}{h}$ steps of diffusion to get an action compared $N$ steps using normal diffusion. However, a smaller number of chunks can still be favorable for policy synthesis because we directly execute the chunk of actions generated in an open loop manner.



\begin{algorithm}
\caption{\modellong Execution}
\label{alg:tedi_sampling}
\begin{algorithmic}[1]
\REQUIRE Buffer of future actions $\mathbf{B}$, Per-action noise level $\mathbf{k}$, Denoiser $\epsilon_\theta(\mathbf{B}, \mathbf{O}_0, \mathbf{k})$,  Observations $\mathbf{O}_t$, Diffusion Noise Levels $N$, Chunk size $T_{B}$, Chunks $h$
\STATE
\STATE $\triangleright$ Initialize buffer $B$ and per action noise levels $\mathbf{k}$
\STATE $\mathbf{B}, \mathbf{k} \gets \text{Initialize buffer}(\mathbf{O}_t)$ 
\STATE
\STATE $\triangleright$ Execute \model in environment.
\WHILE{task not complete} 
\STATE $\triangleright$ Denoise a chunk of actions 
\FOR{$N / h$ denoising iterations} 
    \STATE $\triangleright$ Run one denoising step
    \STATE $\mathbf{B}, \mathbf{k} \gets \epsilon_\theta(\mathbf{B}, \mathbf{O}_t, \mathbf{k})$ \hfill 
\ENDFOR 
\STATE
\STATE $\triangleright$  Execute first action chunk in environment.
\STATE $\mathbf{A}_t \gets \mathbf{B}[:T_b]$ 
\STATE $\text{env}.\text{execute}(\mathbf{A}_t)$
\STATE $\triangleright$  Remove executed action chunk
\STATE $\mathbf{B} \gets \mathbf{B}[T_b:], \mathbf{k} \gets \mathbf{k}[T_b:]$ 

\STATE 
\STATE $\triangleright$  Create new fully noised action chunk
\STATE Sample $\mathbf{z} \sim \mathcal{N}(\mathbf{0}_{T_b}, \mathbf{I}_{T_b})$
\STATE $\triangleright$  Append noise action chunk to buffer / noise levels
\STATE $\mathbf{B}.\text{append}(\mathbf{z}), \mathbf{k}.\text{append}([N]_{\times T_b})$ 
\ENDWHILE 
\end{algorithmic}
\end{algorithm}

\subsubsection{Initializing the action buffer} 
\label{sect:sampling_initial}
 
At the beginning of sampling from \model, the action buffer must be set up for the diffusion steps $k_i$ of chunk $i=1,\ldots,n$. 
We explore three different \textit{action primers} for initializing the action buffer $\mathbf{B}$:
\begin{enumerate}
    \item \textbf{Denoise:} $\mathbf{B} \gets \mathbf{A}_0^{0}$ Where $\mathbf{A}_0^{0}$ is an action sequence obtained by fully denoising every action in trajectory given the initial observation $\mathbf{O}_0$.
    \item \textbf{Constant:} $\mathbf{B} \gets \mathbf{a}_0\text{.repeat}(T_a).$ Assuming that we know a fixed initial action $\rva_0$ to execute, we can repeat this $T_a$ times to form the primer.
    \item \textbf{Zero}: $\mathbf{B} \gets \mathbf{0}. $ Set the action primer to the zero-tensor.
\end{enumerate}
The selected action primer will be added with temporally increasing noise to initialize the action buffer:\begin{equation}\mathbf{B} \gets \text{add\_noise}(\mathbf{B}, \epsilon, \mathbf{k}), \epsilon \sim \mathcal{N}(\mathbf{0}, \mathbf{I}),\label{eq:chunk_wise_noise_levels}\end{equation}where $\text{add\_noise}(\mathbf{B}, \epsilon, \mathbf{k})$ indicates the noising function of the relevant diffusion algorithm. 

%% file: Figures/tedi_sampling_and_noise_injection_schemes.tex
\begin{figure}
    \includegraphics[width=\linewidth]{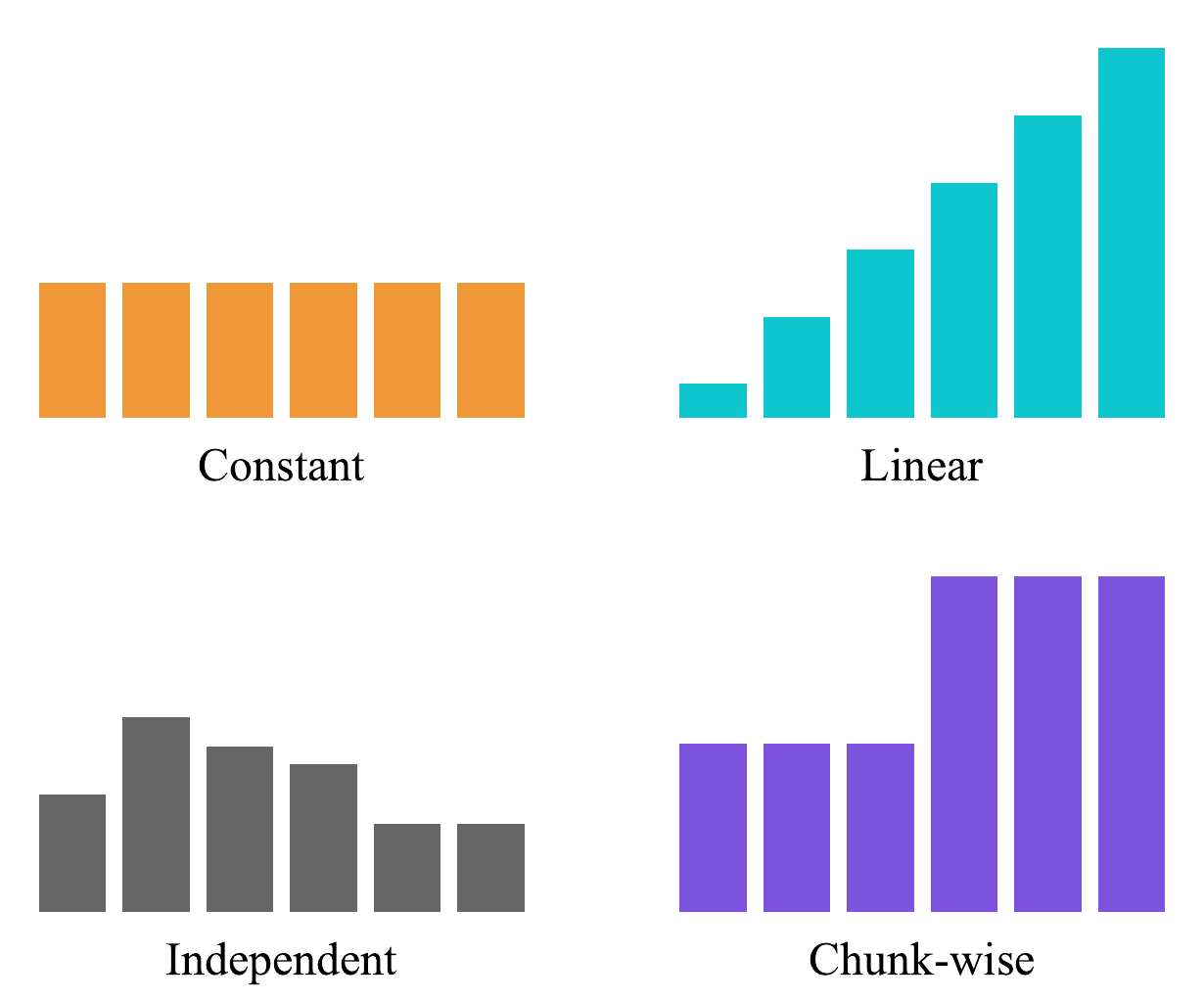}
    \caption{\textbf{Trajectory Noise Corruptions.} Different per-action noise level corruptions applied during training time to train the denoising function in \model.
    }
    \label{fig:noise_injection_schemes}
    \vspace{-20pt}
\end{figure}

%% file: Sections/Method/Subsections/training.tex
\subsection{Training \model Policy}
\label{sec:Training TEDi Policy}

As exemplified by DDPM \cite{hoDenoisingDiffusionProbabilistic2020a}, training a diffusion model generally consists of the following steps: (a) sample a clean data point from the dataset, (b) add noise, (c) have the model predict the added noise or the clean sample. When training a diffusion model on a set of demonstrations, a sample is a clean action sequence $\mathbf{A}_t$, to which noise $\epsilon_k$ with a variance associated with the diffusion level $k$ is added. The sample is associated with an observation sequence $\mathbf{O}_t$, which is interpreted as conditioning for the noise predictor $\epsilon_\theta$, giving rise to the denoising loss
\begin{equation}
    \mathcal{L} = \text{MSE}(\epsilon^k, \epsilon_\theta(\mathbf{A}_t + \epsilon^k ; \mathbf{O}_t, k)).
\end{equation}
In Diffusion Policy, noise is added to the whole sequence with the same variance for each element, as is traditionally done in diffusion models for image generation. With \model, however, we are free to consider a unique variance for each element in $\epsilon^\mathbf{k} = [\epsilon_{k_1},\ldots,\epsilon_{k_h}]^\mathrm{T}$ corresponding to each chunk in the action sequence, resulting in the denoising loss
\begin{equation}
    \mathcal{L} = \text{MSE}(\epsilon^{\mathbf{k}}, \epsilon_\theta(\mathbf{A}_t + \epsilon^{\mathbf{k}}; \mathbf{O}_t, \mathbf{k})), \quad \mathbf{k} \in \mathbb{Z}^{T_p}.
\end{equation}
This opens for different \textit{trajectory noise corruption schemes}, i.e., what $\mathbf{k}=[k_0,\ldots,k_{h-1}]^\mathrm{T}$ is set to during training. We consider the following schemes, which are illustrated in Figure \ref{fig:noise_injection_schemes}:
\begin{enumerate}
	\item \textbf{Constant variance:} The samples are not divided into chunks. Noise with the same variance is applied to all the steps of the action sequence, as in Diffusion Policy.
	\item \textbf{Linearly increasing variance:} The samples are not divided into chunks. Linearly increasing noise is applied to the steps of the action sequence. 
	\item \textbf{Independent:} $\mathbf{k}_i \sim \mathcal{U}[1, \dots, N]$ Each step has independent noise level from a uniform distribution. 
	\item \textbf{Chunk-wise increasing variance:}  Linearly increasing noise is applied to the chunks of the action sequence, where the noise is the same for all steps of a chunk. This noise-corruption scheme will match the denoising steps given to the denoiser during sampling. In particular, for chunk $i$, we a noise levels for the whole chunk ranging from $\mathbf{k}_i \sim \left[\frac{iN}{h}, \frac{(i+1)N}{h}\right]$, corresponding to the noise levels a chunk level denoiser would operate over in Section \ref{sec:Sampling from TEDi Policy}.
 
\end{enumerate}



%% file: Sections/Experiments/experiments.tex
\section{Experiments}
\label{sec:Experiments}

Our experiments answer three questions: (a) \textit{What choice of action primer and noise injection schemes leads to the highest performance?}, (b) \textit{What is the increase in sampling speed for \model relative to standard diffusion-based policies?} and (c) \textit{How does \modellong compare to baseline methods on simulated and real robotic tasks?}

In \ref{sec:buffer_init_expriments}, we set out to find the best-performing combination of action primer and noise corruption scheme to propose a generally performant version of \modellong. Subsection \ref{sec:Sampling Time Experiments} shows the relation between prediction horizon and \model sampling time. Furthermore, subsections \ref{sec:Push-T Evaluation} and \ref{subsec:real_experiments} benchmark \model against relevant baselines on a varied set of simulated and real-world robotic tasks to show that performs comparable to baselines and allows for fast action prediction.


\input{Sections/Experiments/Subsections/buffer_initialization_experiments}

\input{Sections/Experiments/Subsections/sampling_time_experiments}
\input{Sections/Experiments/Subsections/pusht_experiments}

\input{Sections/Experiments/Subsections/RealExperiments}

%% file: Sections/Experiments/Subsections/buffer_initialization_experiments.tex
\input{Sections/Experiments/Subsections/buffer_init_and_noise_injection_results}

\input{Sections/Experiments/Subsections/sampling_time_and_performance_over_Tp_tikz}

\subsection{Action primer and noise corruption scheme}
\label{sec:buffer_init_expriments}
To explore the effect of different parameters for \model, we conduct experiments with the state-based Push-T task, first introduced in \cite{chiDiffusionPolicyVisuomotor2023b}. Here, the policy pushes a T-shaped block lying on a flat surface into a goal area while observing the block's position. The task is lightweight yet complex, requiring the agent to use precise contacts to manipulate the block into the goal area, making it suitable for rapid testing. We base the implementation of \model on the U-Net used in \cite{chiDiffusionPolicyVisuomotor2023b}, only modifying the diffusion step encoder to accept a unique diffusion step for each element in the buffer, not just a single one. Unless specified otherwise, we use 100 training and sampling steps for DDPM and train for 1000 epochs. 

To answer the question of the most performant buffer initialization, we compare the buffer initialization schemes described in Subsection \ref{sec:Sampling from TEDi Policy}; the results are shown in Table \ref{tab:buffer_init_results}. Firstly, we see that the choice of the buffer initialization scheme does not heavily affect performance. We note that the Denoising method requires additional denoising iterations to fully denoise the buffer, resulting in a higher prediction time. Additionally, the Constant method assumes that the actions are observable, namely that $\rva_0$ is part of $\mathbf{O}_0$. While this is true for the Push-T task, it is not generally the case. To achieve a general yet performant algorithm, we choose the Zero method for the rest of our experiments.

We compare the different noise corruption schemes on trajectories in Table \ref{tab:noise_injection_results}, and note that the noise corruption scheme at training highly affects the method's performance, with Linear and Constant performing poorly when used in isolation. 
Prior work in \cite{zhangTEDiTemporallyEntangledDiffusion2023} set a probability of choosing Random at 67\% and 33\% for Linear or set the trajectories with independent levels of noise~\cite{chen2024diffusion}. 
As \model outputs action chunks, not singular actions, we consider a new chunkwise trajectory noise corruption scheme (Section \ref{sec:Training TEDi Policy}). We compare this chunkwise noise corruption scheme with prior noise corruption schemes such as constant, linear and independent. We observe that considering the chunk-wise nature of the sampling is warranted; we achieve a better score by using the chunk-wise noise injection scheme instead of the linear scheme. 
We find that a combination of different schemes performs the best.
Specifically, we find a combination of 80\% Chunk-Wise and 20\% Constant to be the most performant and use this throughout the rest of the experiments.

%% file: Sections/Experiments/Subsections/buffer_init_and_noise_injection_results.tex
\begin{table}[t]
    \centering
    \hfill
    \vspace{0.7em}
    \begin{subtable}[b]{0.49\textwidth}
        \centering
        \begin{tabular}{ccc}
        \toprule
        \textbf{Method} & \textbf{Avg. score} & \textbf{Avg. pred. time}\\
        \midrule
        Zero & 0.90 & 0.30 \\
        Constant & 0.90 & 0.30\\
        Denoising & 0.90 & 0.58\\
        \bottomrule
        \end{tabular}
        \caption{Action primers}
        \label{tab:buffer_init_results}
    \end{subtable}
    \hfill
    \vspace{0.001em}
    \begin{subtable}[b]{0.49\textwidth}
        \centering
        \begin{tabular}{cc}
        \toprule
        \textbf{Noise Corruption Scheme} & \textbf{Avg. score} \\
        \midrule
        Chunk-wise (CW) & 0.86 \\
        Linear & 0.36 \\
        Independent & 0.85 \\
        Constant & 0.29 \\
        67\% Independent 33\% Linear & 0.87 \\
        67\% Independent 33\% CW & 0.89 \\
        80\% CW 20\% Constant & \textbf{0.90} \\
        \bottomrule
        \end{tabular}
        \caption{Trajectory Noise Corruption Schemes}
        \label{tab:noise_injection_results}
    \end{subtable}
    \captionof{table}{\textbf{Abalations.} Comparison of design choices in training and initializing \model on the state-based Push-T task. \textbf{(a) Action primers.} In addition to task performance, we measure the prediction time for each configuration. \textbf{(b) Noise corruption schemes.} We evaluate the proposed trajectory noise corruption schemes along with combinations between schemes.}
    \label{tab:combined_results}
    \vspace{-20pt}
\end{table}

%% file: Sections/Experiments/Subsections/sampling_time_and_performance_over_Tp_tikz.tex

\begin{figure*}[h]
\hfill
\vspace{0.2em}
\begin{minipage}{\textwidth}
    \begin{tikzpicture}
    \definecolor{carrot}{RGB}{240,152,55}
    \definecolor{robinegg}{RGB}{13,197,205}
    \begin{axis}[
        title={Comparison of Prediction Time},
        xlabel={Prediction horizon $T_a$},
        ylabel={Prediction time [s]},
        width=0.5\textwidth,
        height=6cm,
        xmin=7, xmax=47,
        ymin=0.1, ymax=1.2,
        xtick={7, 11, 15, 19, 23, 27, 31, 35, 39, 43, 47},
        ytick={0.1, 0.2, 0.4, 0.6, 0.8, 1.0, 1.2},
        ymajorgrids=true,
        grid style=dashed,
        tick label style={font=\small},
        label style={font=\small},
        title style={font=\small},
        legend style={at={(0.59,0.75)},anchor=west,font=\footnotesize}
    ]
    
    \addplot[
        color=carrot,
        mark=square,
        thick
    ]
    coordinates {
        (7, 1.0868971006944776)(11, 1.1100214791018517)(15, 1.1066246343776585)(19, 1.1316033414099365)(23, 1.157453018380329)(27, 1.1604763683862984)(31, 1.1476071682758628)(35, 1.1526452907826752)(39, 1.1551404526922853)(43, 1.1662846270017326)(47, 1.1558541133068503)
    };
    
    \addplot[
        color=robinegg,
        mark=o,
        thick,
    ]
    coordinates {
        (7, 1.025146345468238)(11, 0.5286022895947099)(15, 0.3476116380188614)(19, 0.3554080847185105)(23, 0.27237563370727)(27, 0.21853902242146434)(31, 0.17991206450387837)(35, 0.1551448745187372)(39, 0.15623727571219206)(43, 0.13883219570852817)(47, 0.12139165899716317)
    };
    \legend{Diffusion Policy,SDP}
    \end{axis}
    \end{tikzpicture}
    \hfill
    \begin{tikzpicture}
    \definecolor{carrot}{RGB}{240,152,55}
    \definecolor{robinegg}{RGB}{13,197,205}
    \begin{axis}[
        title={Prediction Horizon Sensitivity Analysis},
        xlabel={Prediction horizon $T_a$},
        ylabel={Average score},
        width=0.50\textwidth,
        height=6cm,
        xmin=11, xmax=47,
        ymin=0.8, ymax=0.92,
        xtick={11, 15, 19, 23, 27, 31, 35, 39, 43, 47},
        ytick={0.8, 0.82, 0.84, 0.86, 0.88, 0.90, 0.92},
        ymajorgrids=true,
        grid style=dashed,
        tick label style={font=\small},
        label style={font=\small},
        title style={font=\small},
        legend style={at={(0.98,0.98)},anchor=north east,font=\footnotesize}
    ]
    
    \addplot[
        color=carrot,
        mark=square,
        thick,
    ]
    coordinates {
        (11,0.9023)(15,0.9187749094239139)(19,0.8958408361549446)(23,0.8182)(27,0.8712)(31,0.8920151720296038)(35,0.8655189367938235)(39,0.8537)(43,0.8396497777815118)(47,0.803)
    };
    
    \addplot[
        color=robinegg,
        mark=o,
        thick,
    ]
    coordinates {
        (11,0.904)(15,0.9000226166245788)(19,0.8958408361549446)(23,0.891327884222977)(27,0.8581759027233243)(31,0.8502942675086452)(35,0.8538)(39,0.8632541229818115)(43,0.8222)(47,0.8261092849437839)
    };
    \legend{Diffusion Policy,SDP}
    \end{axis}
    \end{tikzpicture}
    \vspace{-15pt}
    \end{minipage}
    \caption{\textbf{Fast Prediction Time with Longer Buffer.} \modellong decreases its sampling time drastically with longer prediction horizons while not sacrificing performance. Here, the chunk length is set to $5$.}
    \label{fig:combined_plots}
    \vspace{-10pt}
\end{figure*}

%% file: Sections/Experiments/Subsections/sampling_time_experiments.tex
\subsection{Sampling Time Experiments}
\label{sec:Sampling Time Experiments}

We measure the relative increase in prediction speed achieved by \model compared to traditional diffusion-based policies. As our baseline, we choose Diffusion Policy~\cite{chiDiffusionPolicyVisuomotor2023b}. Using DDPM for both methods, we measure the prediction time when increasing the horizon, allowing for more chunks of fixed size in the buffer. The results are shown in Figure \ref{fig:combined_plots}.

We see that the prediction time for \modellong decreases monotonically as the buffer grows larger while remaining constant for Diffusion Policy. The decrease in sampling time is due to the design of the sampling scheme introduced in Section \ref{sec:Sampling from TEDi Policy}, where the number of denoising steps required to obtain a clean action chunk is given by $\frac{N}{h}$, where $h$ is the number of chunks in the buffer. We confirm this with our experiment; the sampling time is inversely proportional to the number of chunks. We highlight that this relative speedup extends to other underlying diffusion models, such as DDIM~\cite{songDenoisingDiffusionImplicit2020}.

%% file: Sections/Experiments/Subsections/pusht_experiments.tex
\subsection{Performance on Simulated Robotic Tasks}
\label{sec:Push-T Evaluation}

\begin{figure}
        \centering
        \includegraphics[width=0.48\textwidth]{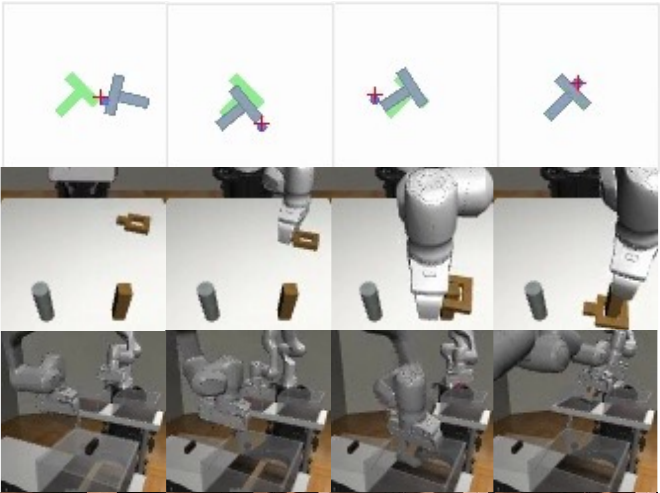}
    \caption{\textbf{Qualitative Illustration of Simulated Experiments.} \model is able to successfully solve complex tasks from image observations, such as Push-T~\cite{chiDiffusionPolicyVisuomotor2023b} and Robomimic-tasks~\cite{pmlr-v164-mandlekar22a}.}
    \label{fig:enter-label}
    \vspace{-15pt}
\end{figure}

\begin{table}
    \centering
    \begin{tabular}{cccc}
    \toprule
    Task & Diffusion Policy & Consistency Policy 3-step & \model \\
    \midrule
    Push-T & \textbf{0.88/0.84} & 0.75/0.68 & 0.84/0.79 \\
    Lift & \textbf{1.00/1.00} & \textbf{1.00/1.00} & \textbf{1.00/1.00} \\
    Can & 1.00/0.98 & 0.97/0.94 & \textbf{1.00/0.99} \\
    Square & 0.98/\textbf{0.93} & 0.95/0.88 & \textbf{0.99/0.93} \\
    Transport & \textbf{1.00/0.88} & 0.89/0.71 & 0.97/\textbf{0.88} \\
    \bottomrule
\end{tabular}
    \caption{\textbf{Simulated Quantitative Results.} (Max, average performance) of \model, Consistency Policy, and Diffusion Policy on Image-based benchmarks across 10 checkpoints over 3 seeds.}
    \label{tab:image_policies}
    \vspace{-15pt}
\end{table}


In this section, we show how \modellong compares to state-of-the-art diffusion-based policies. 
As our baselines, we use Diffusion Policy with settings closely resembling that of \model to show the effects of introducing the novel sampling scheme. In addition, we compare \model with Consistency Policy~\cite{prasadConsistencyPolicyAccelerated2024}, a diffusion-based policy that achieves high sampling speed through Consistency distillation~\cite{kim2024consistency}.

We evaluate all methods on several simulated tasks, including Robomimic-tasks \cite{pmlr-v164-mandlekar22a}, and Push-T \cite{chiDiffusionPolicyVisuomotor2023b}. We use the realistic image-based variant for all tasks, meaning the policy only observes images and proprioceptive sensing. Push-T policies are trained for 1000 epochs. Robomimic Lift, Can, and Square policies are trained for 1500 epochs, while Transport and Tool-hang are trained for 500 epochs. For Consistency Policy, we train both the teacher and the student for the same number of epochs as other methods, doubling the training time compared to the other methods. For Diffusion Policy and Consistency Policy, we set the chunking length to $8$ and the prediction horizon to $15$. For \model, however, we increase the horizon to 19, fitting two chunks in the buffer, leading to a speed-up in the sampling from $2.4$s to $1.8$s. 



The performance of \model is comparable to Diffusion Policy; see Table \ref{tab:image_policies}. This also holds for more intricate tasks, such as Robomimic's Transport, where the policy is tasked with handing over a hammer from one manipulator to another. This shows that the sampling scheme used in \model does not affect performance negatively. Consistency Policy is invariably equal or less performant than the two other methods, which is exaggerated in the more intricate tasks, even though it requires twice the training time.

%% file: Sections/Experiments/Subsections/RealExperiments.tex
\subsection{Real-world experiment}
\label{subsec:real_experiments}

To confirm that \modellong performs well in real-world settings, we benchmark all methods on a real-world Push-T task. We collect 134 demonstrations and train the methods on the dataset for 600 epochs. As in the simulated experiment, we train both the Consistency Policy teacher and student for 600 epochs. The policies are rolled out 20 times during evaluation. We reset the T-block with a script using a fixed random seed, ensuring the same initialization for all methods. We use the DDIM scheduler with 16 steps for Diffusion Policy and \model, and keep two chunks in the buffer of \model.

Due to the design of \modellong, the de-noising of the next chunk has to follow the enactment of the entire previous chunk, and we, therefore, roll out \model synchronously, halting the robot movement while sampling. The motion remains smooth, however, due to the short sampling time of \model. To ensure a fair comparison, we deploy all methods in this fashion.

We report results in Fig. \ref{fig:real_pusht_performance}. We see that SDP achieves the highest score while having a high sampling speed. We observe that the most common failure mode for diffusion policy and consistency policy is that they get stuck performing small oscillatory movements when attempting precise and slow pushes of the block, due to constant action replanning over actions trajectories. This is exacerbated because the demonstrations include many corrective actions as we constantly aim for a perfect end result when demonstrating the task. In contrast, \model, has a persistent action buffer across observations and is less prone to this error. 

\begin{figure}[t]
    \centering
    \hfill
    \vspace{0.5em}
    \begin{subfigure}[b]{0.5\textwidth}
        \centering
        \includegraphics[width=0.7\textwidth]{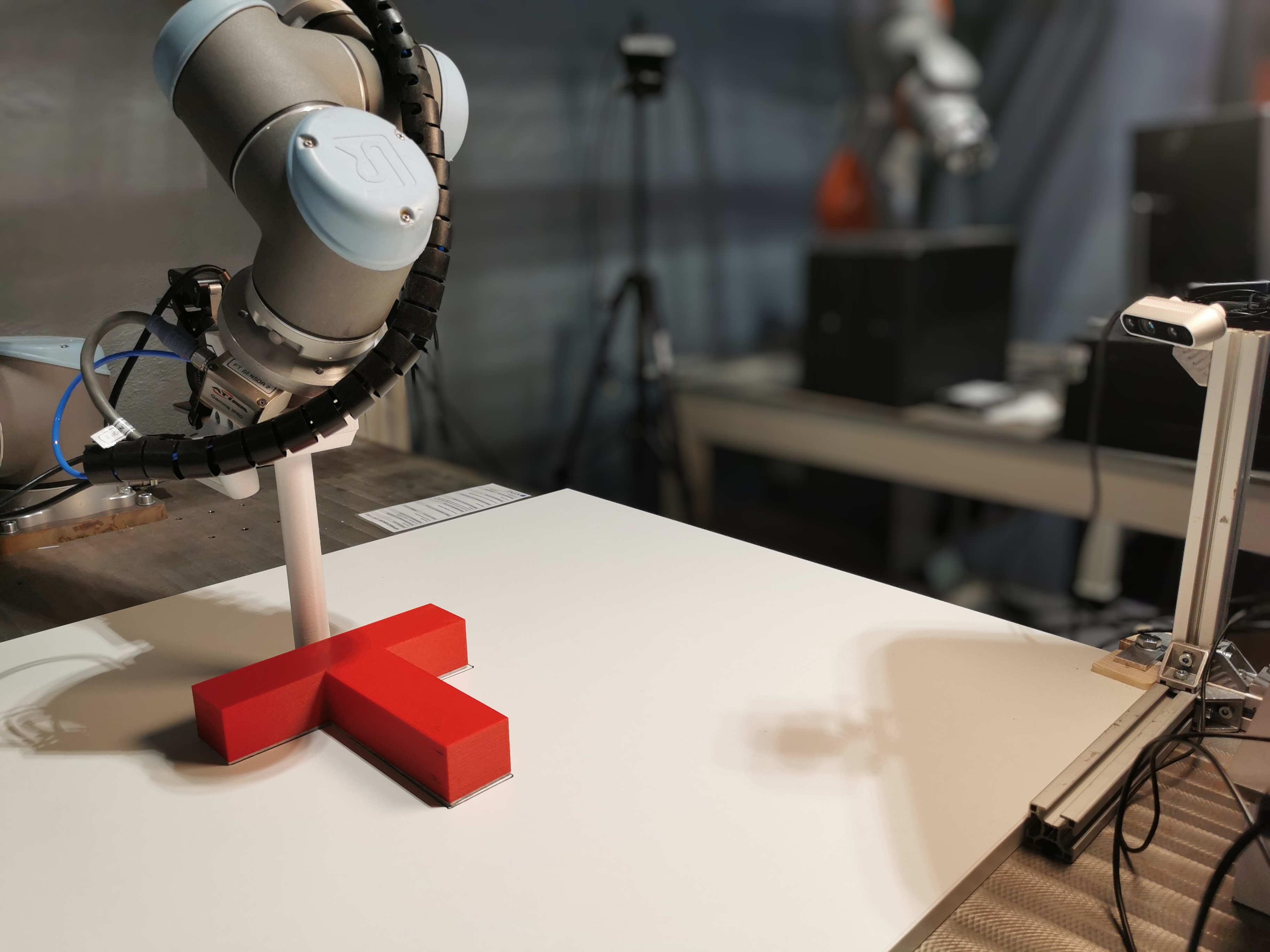}
    \end{subfigure}
    \hfill
    \vspace{.1em}
    \begin{subfigure}[b]{0.5\textwidth}
        \centering
        \includegraphics[width=0.9\textwidth]{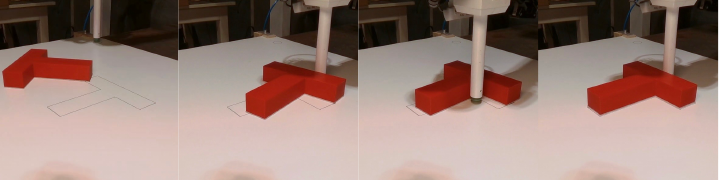}
    \end{subfigure}
    \caption{\textbf{Real-world Push-T Experiment} Hardware setup and task execution illustration. \model is able to precisely push the T-shaped block into the target region.}
    \label{fig:real_pusht_setup}
\end{figure}

\begin{figure}[t]
    \centering
    \begin{subfigure}[b]{0.5\textwidth}
        \centering
        \resizebox{0.9\columnwidth}{!}{
        \begin{tabular}{cccc}
        \toprule
        \textbf{Method} & \textbf{Coverage success} & \textbf{Sampling time}\\
        \midrule
        Diffusion Policy & 0.50  & 0.13s \\
        Consistency Policy & 0.40  & \textbf{0.01s} \\
        \model (Ours) & \textbf{0.85} & 0.07s\\
        \bottomrule
        \end{tabular}
        }
        \label{tab:performance}
    \end{subfigure}
    \caption{\textbf{Real World Push-T Quantitative Results.} At the last step of each episode, we measure the coverage of the block over the goal area. We report the success rate, where success is defined as a higher coverage than the minimum in the demonstration dataset. We also report the average time each method uses to predict an action chunk.}
    \label{fig:real_pusht_performance}
    \vspace{-15pt}
\end{figure}

%% file: Sections/conclusion.tex
\section{Conclusion}
\label{sec:Conclusion}
In this work, we introduce \modellong, a diffusion-based closed-loop policy that drastically reduces the sampling time compared to other diffusion-based policies, such as Diffusion Policy \cite{chiDiffusionPolicyVisuomotor2023b} while achieving high performance on robotic benchmarks. We systematically compare different schemes for noise corruption during training and buffer initialization for sampling. We verify our method on simulated and real-world robotic tasks. 

\paragraph{Limitations and further research.} \model presents a large space of possible configurations, including the choice of chunking length, buffer length and action primer. While we choose a combination that performs well on most tasks, further explorations of combinations might reveal other combinations that perform better on specific tasks. In addition, \modellong introduces a tradeoff, as a longer buffer will increase sampling speed but an excessively long buffer negatively affects performance. Tasks where a reactive policy is needed, might call for a shorter buffer length. In addition, while \model is more robust than the baselines against repeating oscillatory motions during rollouts, alleviating this completely is an interesting topic for further research.
